# Wearable Assistive Devices for the Blind

Ramiro Velázquez

Universidad Panamericana, Aguascalientes, Mexico

**Abstract** Assistive devices are a key aspect in wearable systems for biomedical applications, as they represent potential aids for people with physical and sensory disabilities that might lead to improvements in the quality of life. This chapter focuses on wearable assistive devices for the blind. It intends to review the most significant work done in this area, to present the latest approaches for assisting this population and to understand universal design concepts for the development of wearable assistive devices and systems for the blind.

## 1 Introduction

### 1.1 Target population

Globally, an estimated 40 to 45 million people are totally blind, 135 million have low vision and 314 million have some kind of visual impairment [1]. The incidence and demographics of blindness vary greatly in different parts of the world. In most industrialized countries, approximately 0.4% of the population is blind while in developing countries it rises to 1%. It is estimated by the World Health Organization (WHO) that 87% of the world's blind live in developing countries.

Over the last decades, visual impairment and blindness caused by infectious diseases have been greatly reduced (an indication of the success of international public health action), but there is a visible increase in the number of people who are blind or visually impaired from conditions related to longer life expectancies. The great majority of visually impaired people are aged 65 years or older. It is estimated that there is a per-decade increase of up to 2 million persons over 65 years with visual impairments. This group is growing faster than the overall population.

In younger groups, blindness and visual impairment is mainly due to birth defects and uncorrected refractive errors. In the first case, most of the causes are in the brain rather than in the eye while in the second one, they are conditions that could have been prevented if diagnosed and corrected with glasses or refractive surgery on time.

It is estimated that by the year 2020, all blind-related numbers will double.

### 1.2 Dimension of the problem

Of all sensations perceived through our senses, those received through sight have by far the greatest influence on perception. Sight combined with the other senses, mainly hearing, allow us to have a world global perception and to perform actions upon it. For the blind, the lack of sight is a major barrier in daily living: information access, mobility, way finding, interaction with the environment and with other people, among others, are challenging issues.

In fact, school and working-age blind have very high analphabet and unemployment rates. For example, in the US, the blind unemployment rate is around 75% while only 10% of the blind children receive instruction in Braille [2]. Despite efforts, a true is that most schools and employers cannot accommodate blind people. In consequence, the person who is blind and his/her family face important socioeconomic constraints.

The issue of the blind becomes a very serious problem in terms of health and social security. Costly-in home expenses, nursing home care and welfare expenses on unemployment and health services have to be absorbed by the state.

A state action to enable the blind/visually impaired to live independent and productive lives has been to teach them new ways to accomplish routine daily tasks. A great variety of specialists is involved: special education teachers, Braille teachers, psychologists, orientation and mobility specialists, low-vision specialists and vision rehabilitation therapists to name a few.

Evidently, this involves a very high cost that has to be absorbed by the state. Moreover, the availability of funding and qualified personnel is insufficient to cover the actual population's demand. Other means are urgently needed to assist this population.

### 1.3 Assistive technology

Advances of technology and better knowledge in human psycho-physiological 3D world perception permit the design and development of new powerful and fast interfaces assisting humans with disabilities. For the blind, research on supportive systems has traditionally focused on two main areas: information transmission and mobility assistance. More recently, computer access has been added to the list [3].

Problems related to information transmission concern reading, character recognition and rendering graphic information about 2D and 3D scenes. The most successful reading tool is the Braille dot code. Introduced by Louis Braille in the 19$^{th}$ century, it has now become a standard worldwide. Inventions addressing the problems of character recognition and pictorial representation mostly consist of tactile displays. They permit character and graphic recognition by feeling a tactile version of them.

Problems related to mobility assistance are more challenging. They involve spatial information of the immediate environment, orientation and obstacle avoidance. Many electronic travel aids (ETAs) for safe and independent mobility of the blind have been proposed over the last decades. They all share the same operation principle: they all scan the environment (using different technologies) and display the information gathered to other sense (mainly hearing and touch).

With the internet revolution of the last years, problems related to computer access for the blind arose. Popular solutions are voice synthesizers, screen magnifiers and Braille output terminals. Voice synthesizers practically read the computer screen; screen magnifiers enable on-screen magnification for those with low-vision and Braille output terminals are plugged to the computer so that information on the screen is displayed in Braille.

### 1.4 Wearable and portable devices

This chapter reviews wearable assistive devices for the blind and less portable assistive devices. There is a slight difference between both.

Wearable devices are distinctive from portable devices by allowing hands-free interaction, or at least minimizing the use of hands when using the device. This is achieved by devices that are actually worn on the body such as head-mounted devices, wristbands, vests, belts, shoes, etc.

Portable devices are usually compact, lightweight, they can be easily carried (but not worn) by the user and require constant hand interaction. For example: tactile displays, electronic canes, mobile phones, laptop computers, etc.

The area of wearable devices is currently a "hot" research topic in assisting people with disabilities such as the blind. As this area is still very much young and experimental, there are not many mature commercial products with a wide user base.

## 2 Wearable technologies for the blind

A number of wearable assistive devices have been developed as task-specific solutions for activities such as reading and travel. Given the fact that sight is missing, they try to open new communication channels through hearing and touch. Devices are as diverse as the technology used and the location on the body. Fig. 1 overviews the body areas involved in wearable assistive devices: fingers, hands, wrist, abdomen, chest, feet, tongue, ears, etc. have been studied to transmit visual information to the blind. This chapter intends to review several prototypes so that their potential can be appreciated.

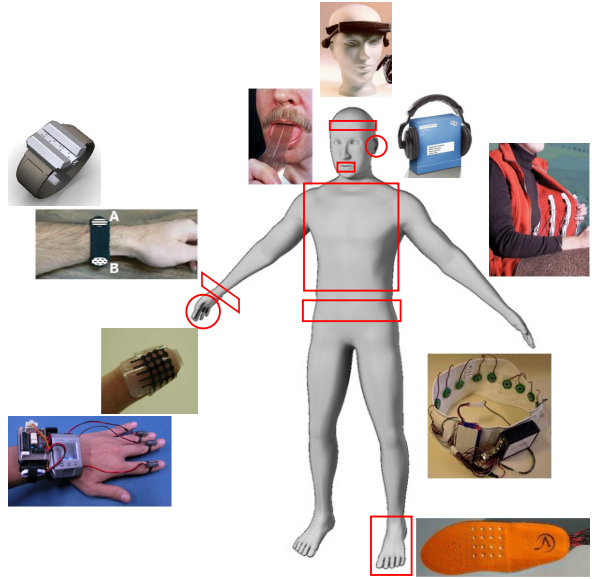

**Fig. 1.** Overview of wearable assistive devices for the blind.

## 2.1 Considerations on hearing and touch

After sight, hearing and touch are definitively the second and third major human senses, respectively. For the blind, they evidently become the first and second ones, respectively.

Blind people rely on hearing environmental cues for key tasks such as: awareness, orientation, mobility and safety. A representative example is when trying to cross a street intersection all alone: they stand-still listening to the environment and will not cross until the traffic light sequence is fully understood.

The ear is the sense organ that detects sound vibrations. It is responsible for transducing these vibrations into nerve impulses that are perceived by the brain. After brain processing, it is possible for humans to detect several characteristics of sound such as loudness, pitch or frequency, timbre, direction and distance to the source.

Roughly, the audible frequency range of the human ear is 20 Hz to 20 kHz with a remarkable discrimination of 0.5 to 1 Hz. Hearing a sound mainly depends on 2 parameters: sound intensity and frequency.

Fig. 2 shows the measured threshold of hearing curve that describes the sound intensity required to be heard according to frequency. The standard threshold of hearing at 1 kHz is nominally taken to be 0 dB, but it is actually about 4 dB. Note that there is a marked difference between low and high frequencies: while about 60 dB is required to be heard at 30 Hz, about 18 dB is required at 10 kHz. The high sensitivity region at 2 to 5 kHz is very important for the understanding of speech.

Human tolerance to sound intensity goes from the threshold of hearing at 4 dB to the threshold of pain at 120 dB, which can be produced by a jet engine.

It has been demonstrated that 8 hours of 90 dB sounds can cause damage to the ears, 1 minute of 110 dB causes hearing loss and any exposure to 140 dB sounds causes immediate and irreversible damage [4].

Additional related problems are the degradation and overload of the hearing sense. Recent studies [5] have shown that a 20-30 minute listening to music/speech/sound activity causes degradation to human sensors information registration, reduces human capacity to perform usual tasks and affects the posture and equilibrium.

All these facts must be taken into account when designing assistive devices that exploit hearing as the substitution sense.

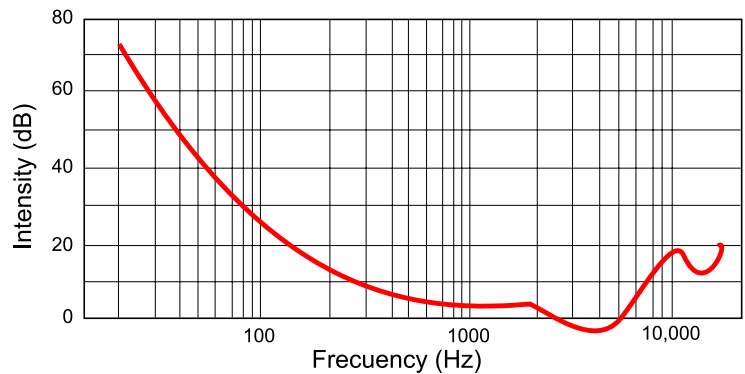

**Fig. 2.** Threshold of hearing: sound intensity minimal thresholds as function of frequency, after [4].

In healthy-sighted, touch is generally used as an additional independent sensory modality to convey information or as a redundant modality to increase information coming from vision and hearing.

For the blind, touch becomes the primary input for the receipt of non-audible physical information. Blind people can rapidly and accurately identify three-dimensional objects by touch. They can also locate and orient themselves in known environments by touching objects. Braille readers access information through touch.

The skin is the sense organ that contains the essential biological sensors of touch. It encompasses 3 main groups of sensors organized by biological function: the thermoreceptors, responsible for thermal sensing, the nociceptors, responsible for pain sensing and the mechanoreceptors, sensitive to mechanical stimulus and skin deformation.

Our interest focuses on the mechanoreceptors as they are responsible for sensing and transmission of physical deformations by external forces to the nervous system. Four kinds of mechanoreceptors can be found on the human glabrous skin: Pacini corpuscles, Ruffini endings, Merkel cells and Meissner corpuscles.

According to [6], Meissner corpuscles respond to touch, Pacini corpuscles respond to vibration, Ruffini endings respond to lateral extension of the skin and articular movement and Merkel cells perceive pressure.

Our interest mainly focuses on Meissner and Pacini skin mechanoreceptors since they are involved in hand feeling during object exploration.

Similar to the relationship found for hearing, feeling a deformation on the skin depends on the relation between the amount of skin indentation and frequency. Fig. 3 shows this relation for Meissner and Pacini mechanoreceptors. Note that while Pacini corpuscles are sensitive to low amplitude-high frequency stimuli, Meissner ones are sensitive to high amplitude-low frequency stimuli.

The ability to discriminate stimuli on the skin varies throughout the body. The two-point discrimination threshold (TPDT) is a measure that represents how far apart two pressure points must be in order to be perceived as two distinct points on the skin [8]. This measurement usually aids designers in choosing the density of a tactile display depending on the part of the body the tactile display will be mounted. Fig. 4 shows the TPDT for different areas of the body. The TPDT is 2.54 mm for the fingertips, 18 mm for the forehead, 40 mm for the forearm, 42 mm for the back, 49 mm for the calf, etc.

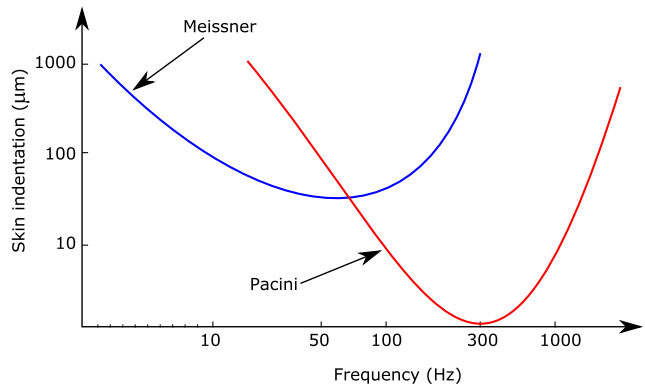

**Fig. 3.** Skin indentation minimal detection thresholds as function of frequency, after [7].

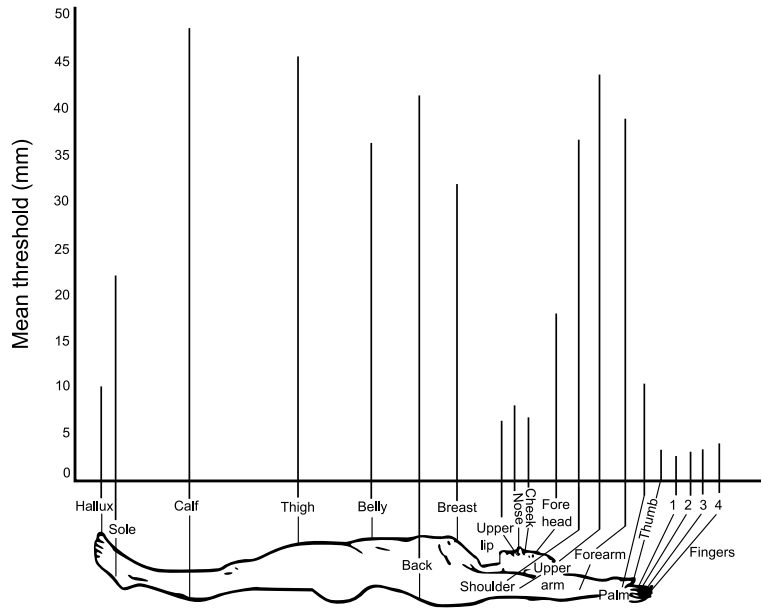

**Fig. 4.** TPDT for different areas of the body, after [9].

Figs. 3 and 4 are the basis for designing wearable touch stimulation devices such as tactile displays. For example, for correctly stimulating the fingertips, a tactile display should integrate an actuator array of 2.54 mm interspacing (Fig. 4) and produce an effective skin indentation of 1 mm at 1 Hz (Meissner curve, Fig. 3) or have an array of contact pins at 2.54 mm interspacing and produce 10 μm of skin indentation at 100 Hz (Pacini curve, Fig. 3).

### 2.2 Assistive devices worn on fingers and hands

Most of the assistive devices for the blind that exploit touch as the substitution sense are tactile displays for the fingertips and palms. Typical tactile displays involve arrays of vibrators or upward/downward moveable pins as skin indentation mechanisms.

Many tactile devices have been developed using a wide range of technologies. Approaches range from traditional actuation technologies such as servomotors, electromagnetic coils, piezoelectric ceramics and pneumatics [10-13] to the new ones: shape memory alloys (SMAs), electroactive polymers (EAPs), electrorheological (ER) fluids and airbone ultrasound [14-17].

However, most of the tactile displays found in the literature are at best good examples of portable devices. Wearable devices for the fingers and palms were not found in the literature until recently. Two examples are the band-aid-size tactile display from Sungkyunkwan University (Korea) and the Finger-Braille interface from Tokyo University (Japan).

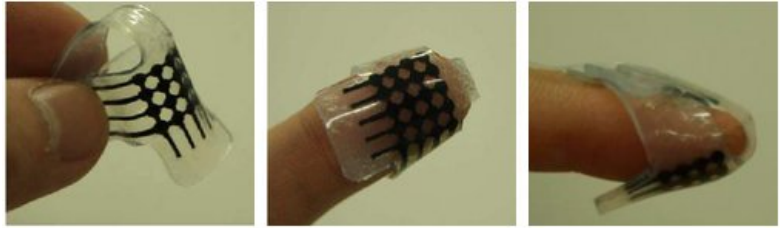

**Fig. 5.** Wearable tactile display for the fingertip.

The bandage-sized tactile display is an innovative touch stimulation device based on EAP soft actuator technology (Fig. 5). It is soft and flexible and can to be wrapped around the finger like a band-aid. This new wearable display could be used as a Braille display or as a multi-purpose tactile display to convey visual information to the blind.

The first prototype developed is a thin polymer sheet of 210 μm thick with 20 EAP soft actuators distributed in an active area of 14 x 11mm$^2$ which covers most of the fingertip's touch-sensitive area. Tactile feel is produced by actuating the 20 contact points independently. Both vibration and upward/downward patterns can be generated using an external user computer interface [18].

The Japanese Finger-Braille interface is a wearable assistive device to communicate information to the deaf-blind. In this system, the fingers are regarded as Braille dots: 6 fingers, 3 at each hand, are enough to code any 6-dot Braille character. Some examples of translation are shown in fig. 6.

Using this codification, 6 small lightweight vibrating DC motors were attached to the fingers (Fig.7). They provide a 120 Hz vibration to stimulate the back of the finger. Each 3 motor-hand is controlled by a Citizen-IBM wristwatch computer and an electronic module that includes the batteries and control circuitry. The wristwatch computer is capable of communicating with external devices via Bluetooth technology. The total weight of the equipment, including battery, is approximately 170 g per hand.

The creators of the Finger-Braille interface report in [19] high recognition rates from experiments conducted with deaf-blind subjects which shows the prototype's potential for providing Braille information.

**Fig. 6.** Finger-Braille code.

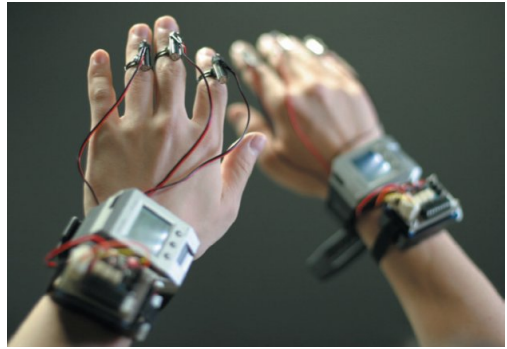

**Fig. 7.** Finger-Braille interface with wristwatch computers.

### 2.3 Assistive devices worn on the wrist and forearm

Researchers at the University of British Columbia (Canada) have developed two prototypes of vibrotactile displays that can be worn on the forearm and wrist (Fig. 8(a)).

Both displays consist of 2 vibrating DC motors spaced 60 mm apart that generate vibrations at 140 Hz [20].

These tactile devices convey information using intermittent alert-like signals and were used to alert clinicians of adverse changes in a patient's heart rate without distracting their attention with auditory alarms. Experiments conducted with these prototypes showed that there was no difference between the wrist and forearm: comfort and accuracy of information were perceived to be the same.

Similarly to this application, these devices could be used to convey simple patterns such as alert-like information to the blind for example when approaching an obstacle.

A popular wearable assistive device for the wrist is of course the Braille watch (Fig. 8(b)). On the basis that a watch is a primary necessity for living a normal life, several companies are nowadays commercializing Braille watches.

These watches, which have the exact same mechanism as the regular ones, display time information as raised patterns on the dial or as Braille numbers which blind people will sense.

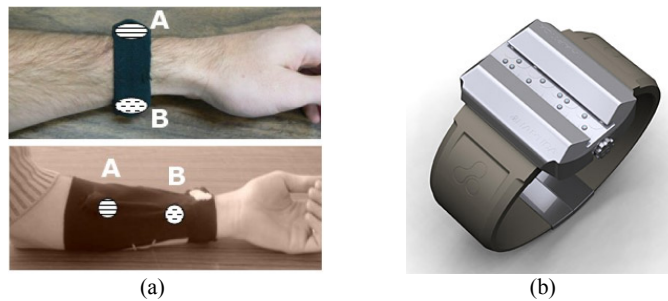

(a)  (b)

**Fig. 8.** (a) Tactile display prototypes for the forearm and wrist and (b) Braille watch.

## 2.4 Assistive devices worn on the tongue

In normal vision, the eyes send signals to the middle of the brain. From there, these signals are sent directly to the visual cortex. Not so for the blind.

In 1998, the University of Wisconsin (USA) introduced the TDU (Tongue Display Unit). The TDU proposed to retrain the way the brain processes visual information by first stimulating the tongue with an electrode array. The nerves in the tongue send signals through a different pathway to the brain stem in the area that deals with touch. Eventually, the blind person learns to interpret touch as sight in the virtual cortex.

The TDU first prototype translated optical images picked up by a camera into electro-tactile stimuli which were delivered to the dorsum of the tongue via a 12 x 12 flexible electrode array placed in the mouth (Fig. 9(a)). Experiments report that after sufficient training (15 h), the user loses awareness of on-the-tongue sensations and perceives the stimulation as shapes and features in space [21].

Inspired by the TDU, researchers at the University of Montreal (Canada) proposed a tongue display to help blind people navigate around obstacles (Fig. 9(b)) [22]. This device, under the name of Brainport Vision Technology, is expected to be commercially available in the near future.

Another TDU-based display is the one developed by researchers at Joseph Fourier University (France). Their prototype consists of a matrix of 36 electrodes that transmit electrical impulses to the tongue (Fig. 9(c)). This device is currently being used to detect and correct stability and posture [23].

The ultimately goal of the TDUs is to develop a compact, cosmetically acceptable, wireless system that can be worn like a dental orthodontic retainer.

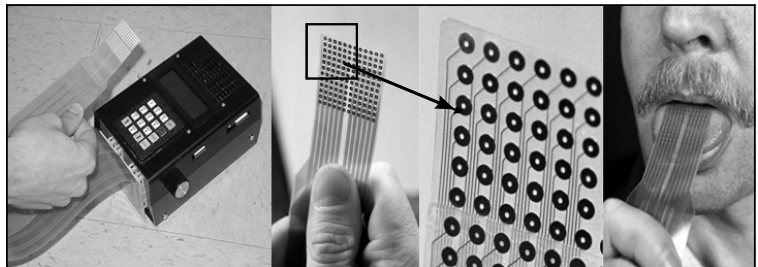

(a)

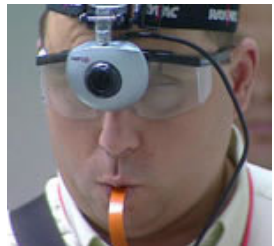 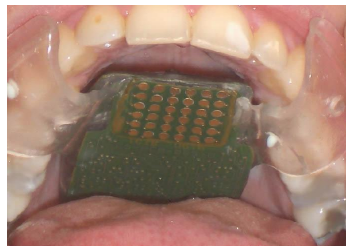

(b)          (c)

**Fig. 9.** TDU systems. Prototypes of: (a) the University of Wisconsin, (b) the University of Montreal and (c) Joseph Fourier University.

## 2.5 Head-mounted assistive devices

Head-mounted devices (HMDs) such as headsets and headbands are the most popular kind of wearable assistive devices.

The head is the rostral part of the human body that comprises the brain, eyes, ears, nose and mouth (all of which are responsible for sensory functions). In particular, the ears, the only sensory organ responsible for hearing throughout the body and the main substitution pathway for the blind, are located on the head. Moreover, humans use head motion to gather information from the environment. It is easy to deduce that devices worn on the head acquire that freedom of motion for environment scanning.

A number of HMDs have been developed for reading and travel assistance of the blind. For reading, the most representative example is the audio book. Entire text books are recorded in the form of speech and reproduced by wearable headset systems (earphones and player).

Since 1950, audio books have benefited the blind population by offering a simple, low-cost, non-Braille reading option. Thousands of titles are available, and with today's technology they can be downloaded for example, from the National Library Service for the Blind and Physically Handicapped (NLS) [24]. Even though the undisputable advantages of audio books, two issues have to be considered:

- **Availability**. Even though there are thousands of titles, not all books are systematically converted to audio books and there is a significant delay with new books. Months, years (or never) could take to have a new book available in audio form.
- **Audio books should not be considered as the reading solution for blind people**. It is true that they fit perfectly for the elderly and non-Braille readers. However as healthy-sighted do prefer to read a book instead of hearing it, why do we assume that a blind person prefers to listen to a book instead of reading it? For the blind, reading a book heightens the self-esteem and independence. Moreover, it trains their orthography which is in most cases quite bad.

For travel assistance, several HMDs have been proposed. In the early years, the most sophisticated device that first became commercially available was the Binaural Sonic Aid (SonicGuide) [25]. The SonicGuide consisted of an ultrasonic wide-beam equipment mounted on spectacle lenses (Fig. 10(a)). Signals reflected back from the 3D world were presented to the user as audio indicating the presence of an obstacle and its approximate distance to the user.

Over the years, the SonicGuide has undergone continuous improvements and its latest version is the system called KASPA [26]. KASPA is worn as a headband (Fig. 10(b)) and creates an auditory representation of the objects ahead of the user. With sufficient training, allows users to distinguish different objects and even different surfaces in the environment.

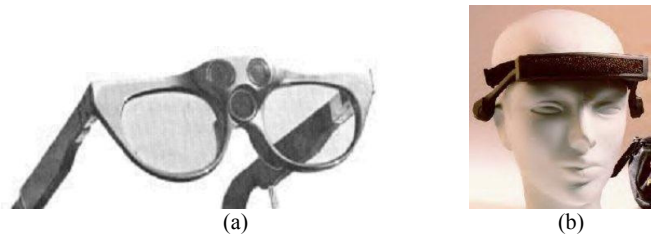

(a)          (b)

**Fig. 10.** (a) The SonicGuide (1974) and (b) its latest version, the KASPA system (2002).

Many portable ETAs like the SonicGuide and the Kaspa system have been developed in the form of hand held devices [27-28]. All these devices are similar to radar systems: a laser or ultrasonic beam is emitted in a certain direction in space and the beam is reflected back from objects that it confronts on its way. A sensor detects the reflected beam, measures the distance to the object and indicates that information to the user through audio or tactile signals.

A new generation of ETAs aims to provide a sensory substitution/supplementation more than merely obstacle detection. The information is acquired using video cameras and its processing is more at cognitive level (brain plasticity, perception, intentionality, etc.). Two representative examples are here presented.

The vOICe system (Fig. 11(a)), patented by Philips Co., converts visual depth information into an auditory representation [29]. This is the first device that successfully uses a camera as an input source. An image is translated to sounds where frequency and loudness represents different scene information parameters such as position, elevation and brightness (Fig. 11(b)). Simple things like finding an object may be mastered in minutes but fully mastering the vOICe's visual-to-auditory language may well take years.

The vOICe is nowadays a mature commercial product [30] and a large number of testimonials show that it is actually improving the quality of life of blind users.

The Intelligent Glasses (IG) is a combined HMD and tactile display system developed at Paris 6 University (France). The IG is a new generation ETA that provides tactile maps of visual spaces and allows users to deduce possible paths for navigating these spaces in order to perform independent, safe and efficient mobility tasks.

The IG system is basically composed of 3 main modules: vision module, scene analyzer and tactile display. Fig. 12 shows the IG first wearable prototype and its operation principle: (a) a pair of stereo-cameras mounted on the glasses frame acquire the environment's representation. (b) Vision algorithms are then applied in order to identify the obstacles in the scene and their user-related position. (c) Finally, this information is displayed on a tactile display for fast exploration by the user. The resulting tactile map is a simple edge-like representation of the obstacles' locations in the scene. All obstacles are considered and displayed in tactile domain as binary data: presence or absence of an obstacle.

Results in [31] show that healthy-sighted blindfolded subjects are able to understand, interact and navigate spaces using tactile maps. Upon training, subjects become more efficient and used to the IG system and tactile maps.

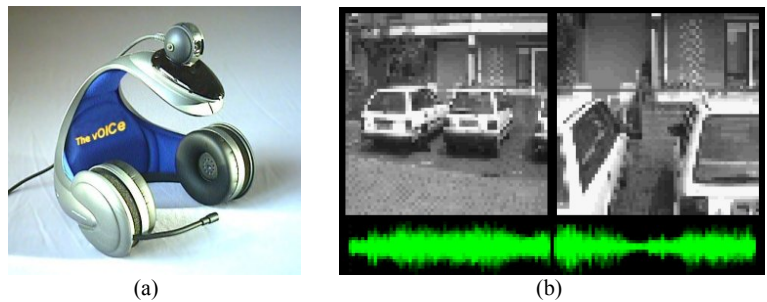

**Fig. 11.** (a) The vOICe system and (b) an example of its image-to-sound rendering.

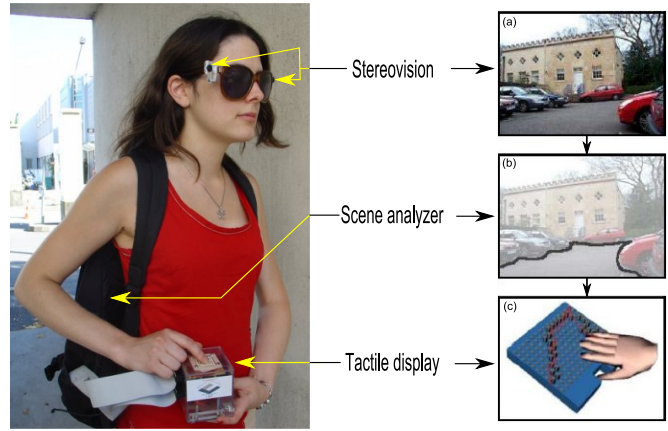

**Fig. 12.** The IG wearable system and an example of its image-to-tactile rendering.

### 2.6 Vests and belts

Researchers at Carnegie Mellon University (USA) presented in [32] the Kahru Tactile Outdoor Navigator (Fig. 13(a)). The Kahru system is a wearable tactile harness-vest display that provides simple directional navigation instructions. A set of 6 vibrating motors generates tactile messages such as forward, back, left, right, speed up and slow down to guide the user through an environment. Communication with the vest is ensured by a belt-worn infrared receiver.

TNO Human Factors, an applied scientific research institute in the Netherlands, has developed a tactile display that consists of 128 vibrating elements attached to a vest (Fig. 13(b)). Vibrations at 160 Hz present 3D spatial information to the user. This vest is currently being used to convey flight information to pilots in an intuitive way [33]. Similarly, it could be used for the blind.

Researchers at MIT (USA) have developed a tactile display embedded in a vest that fastens around the lower torso (Fig. 13(c)). This tactile display consists of a 4 x 4 array of vibrating motors which are independently controlled by an electronic unit. The electronic unit receives commands wirelessly from a remote computer.

This tactile vest display can be used as a navigation aid outdoors, as experiments in [34] have proved that 8 different vibrotactile patterns can be interpreted as directional (for example: stop, look left, run, proceed faster or proceed slower) or instructional cues (for example: raise arm horizontally, raise arm vertically) with almost perfect accuracy.

The NavBelt [35], a wearable ETA proposed by the University of Michigan (USA), provides acoustical feedback from an array of ultrasonic sensors mounted on a belt around the abdomen (Fig. 14(a)). These sensors provide information on very local obstacles placed in a 120° wide sector ahead of the user.

Researchers at Keio University (Japan) proposed in [36] the ActiveBelt, a belt-type tactile display for directional navigation. The ActiveBelt consists of a GPS, a geomagnetic sensor and 8 vibrators distributed at regular intervals around the torso (Fig. 14(b)).

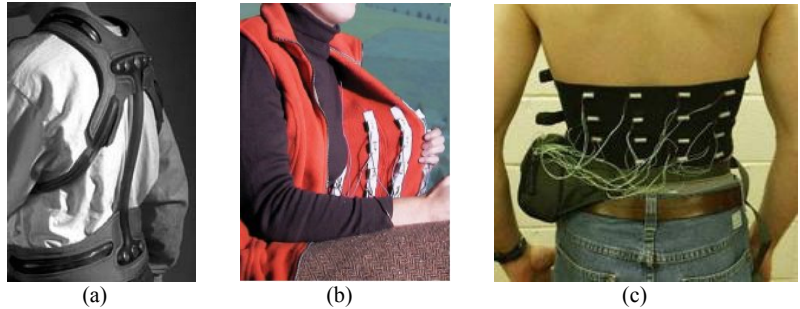

**Fig. 13.** Tactile vest displays. Prototypes of: (a) Carnegie Mellon University, (b) TNO Human Factors and (c) MIT.

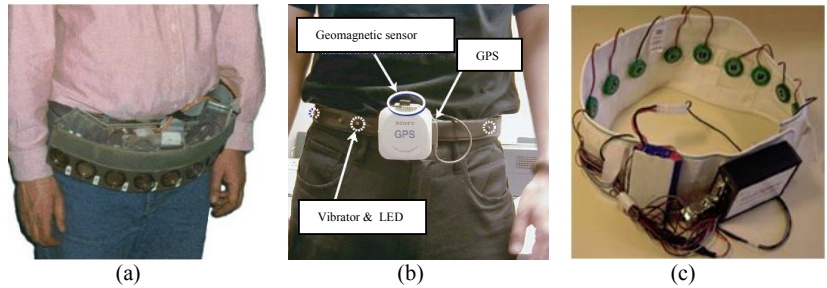

**Fig. 14.** Belt-worn assistive devices. Prototypes of: (a) the University of Michigan, (b) Keio University and (c) the University of Osnabrück.

Vibrations within the range of 33 to 77 Hz are supplied by the ActiveBelt to indicate directions to the user. A set of experiments confirmed that subjects are able to identify the 8 directions while walking.

Another belt-type assistive device is the one developed at the University of Osnabrück (Germany). The prototype consists of an electronic compass and 13 vibrators located around the abdomen (Fig. 14(c)). This belt enables its user to continuously feel his orientation in space via vibrotactile stimulation. Navigation accuracy and long-term usage of the belt are currently being evaluated [37].

### 2.7 Assistive devices worn on the feet

The human foot is a highly functional structure yet its full capabilities have not been thoroughly explored.

What we know about the human foot is that it combines mechanical complexity and structural strength. The ankle serves as foundation, shock absorber and propulsion engine. The foot can sustain enormous pressure and provides flexibility and resiliency. Sensory input from the foot, particularly from the foot sole, has long been recognized as an important source of sensory information in controlling movement and standing balance [38]. As the load on the foot is transferred from heel to toe, pressure signals are automatically fed back to the brain to provide important information about the body's position with respect to the supporting surface.

Our work at Panamericana University (Mexico) has focused on evaluating the performance of the foot sole receptors for information transmission.

For this purpose, we have developed a shoe-integrated vibrotactile display to study how people understand information through their feet and to evaluate whether or not this comprehension level is sufficient to be exploited for assistance of the blind.

Based on the physiology of the plantar surface of the foot, a first prototype consisting of a 16-point array of actuators has been designed and implemented (Fig. 15). All 16 vibrators have been successfully integrated in a regular foam shoe-insole with 10 mm interspacing. They provide vibrating frequencies between 10-55 Hz. Each vibrator is independently controlled with a specific vibrating frequency command.

One of the advantages of this mechatronic shoe-insole is that it can be further inserted into a shoe making it an inconspicuous and visually unnoticeable assistive device. Unlike other portable/wearable assistive devices, an on-shoe device does not heighten the handicapped image that affects the user's self-esteem.

Experiments involving direction, shape, pattern recognition and navigation in space have been conducted with healthy blindfolded-sighted and blind people to gain insights into the capabilities of tactile-foot perception [39].

Results show that both healthy-sighted and blind subjects understand easily vibrations encoding simple information such as directional instructions (for example: go forward, backward, turn left, turn right and stop) and familiar patterns (for example: SMS, phone call, caution) but do not understand vibrations encoding more complex information such as shapes. Although it seems that the feet are not appropriate for precise information recognition, collected data show that it is feasible to exploit podotactile information for navigation in space.

Current work evaluates (1) whether long-term vibrating stimuli on the foot affects balance and walking and (2) user performance depending on cognitive load. The final goal is to integrate the concept of podotactile stimulation in ETAs for the blind.

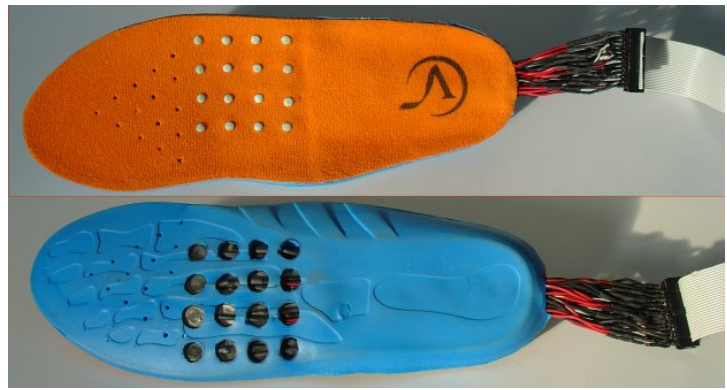

**Fig. 15.** Shoe-integrated tactile display: back and forth. Prototype of Panamericana University.

## 3 Synthesis and conclusions

The miniaturization of actuators and electronics has allowed the creation of new devices and systems that can be embedded into clothing. These wearable systems facilitate the user's ability to perform normal daily tasks without feeling encumbered by burdensome devices.

In particular, this chapter has focused on wearable assistive devices for the blind. A brief non-exhaustive survey of wearable assistive devices for this population has been presented to illustrate the most representative work done in this area. Devices worn on the finger, hands, wrist, forearm, tongue, head, chest, abdomen and feet have been proposed over the last decades to provide wearable solutions to the problems of reading and mobility.

For the blind, hearing and touch become the first and second major senses, respectively. They will never replace vision but they still gather much information from the environment for daily tasks. That is the reason why assistive devices provide acoustical and tactile feedback to compensate for visual information. In contrast, smell and taste are largely ignored as being essential to the interaction with the environment.

Several universal design concepts for acoustical/tactile based assistive devices have been presented. They provide guidelines to stimulate both hearing and touch in order to obtain the best performance from these senses. Yet, some considerations must be taken into account:

- **Sensory overload**. The brain simultaneously processes stimuli from several or all of the sensory modalities to interpret the surrounding environment. Because humans have a limited capacity to receive, hold in working memory and cognitively process information taken from the environment, the use of only one sensory modality to convey information can quickly overload that modality. After a while, users may be limited in the perception of acoustical or tactile signals coming from assistive devices.
- **Long learning/training time**. Learning and mastering the visual-to-sound or visual-to-tactile new language is quite a challenge and requires long training time, patience and great effort from the user.
- **Acoustical feedback** is useful only for reading applications. For mobility applications, it might interfere with the blind person's ability to pick up environmental cues. Moreover, continuous acoustic feedback (20-30 min) might affect posture and equilibrium.
- **Tactile feedback** is a viable choice for mobility applications. However, the information presented must be in accordance with the location of the tactile display on the body. Precise information can only be recognized with the fingers and tongue while simple information can be displayed on the rest of the body. It seems that simple directional information is the best choice for mobility of the blind. It does not require constant activity and cognitive effort that reduces walking speed and quickly fatigues the user.

Despite efforts and the great variety of wearable assistive devices available, user acceptance is quite low. Audio books and Braille displays (for those who can read Braille) and the white cane and guide dog will continue to be the most popular reading/travel assistive devices for the blind.

Acceptance of any other portable or wearable assistive device is always a challenge in blind population. Motivation, cooperation, optimism, willingness/ability to learn or adapt new skills is not a combination that can be taken for granted.